\documentclass[letterpaper,journal]{IEEEtran}
\usepackage{amsmath,amsfonts}
\usepackage{array}
\usepackage{textcomp}
\usepackage{stfloats}
\usepackage{url}
\usepackage{verbatim}
\usepackage{graphicx}
\usepackage{cite}
\usepackage{xcolor}
\usepackage{subcaption}
\usepackage{mathtools}  
\usepackage{amssymb}
\usepackage{tabulary}
\usepackage{booktabs}
\usepackage[ruled,linesnumbered]{algorithm2e}
\usepackage{hyperref}
\usepackage{setspace}

\newcommand{\sss}{\scriptscriptstyle}
\newcommand{\uvec}[1]{\overrightarrow{#1}}

\title{
CoNi-MPC: Cooperative Non-inertial Frame Based Model Predictive Control
}

\author{
        \IEEEauthorblockN{
        Baozhe Zhang\IEEEauthorrefmark{2}\textsuperscript{1, 3}, 
        Xinwei Chen\IEEEauthorrefmark{2}\textsuperscript{1, 2}, 
        Zhehan Li\textsuperscript{1, 2}, \\
        Giovanni Beltrame\textsuperscript{4},
        Chao Xu\textsuperscript{1, 2},
        Fei Gao\textsuperscript{1, 2},
        and Yanjun Cao\textsuperscript{1, 2} 
        }
        \vspace{-1.0cm}
        }

\begin{document}

\twocolumn[{
\renewcommand\twocolumn[1][]{#1}
\maketitle
\begin{center}
\setlength{\abovecaptionskip}{-0.10cm}
\setlength{\belowcaptionskip}{0.25cm}
\centering
\includegraphics[width=\textwidth]{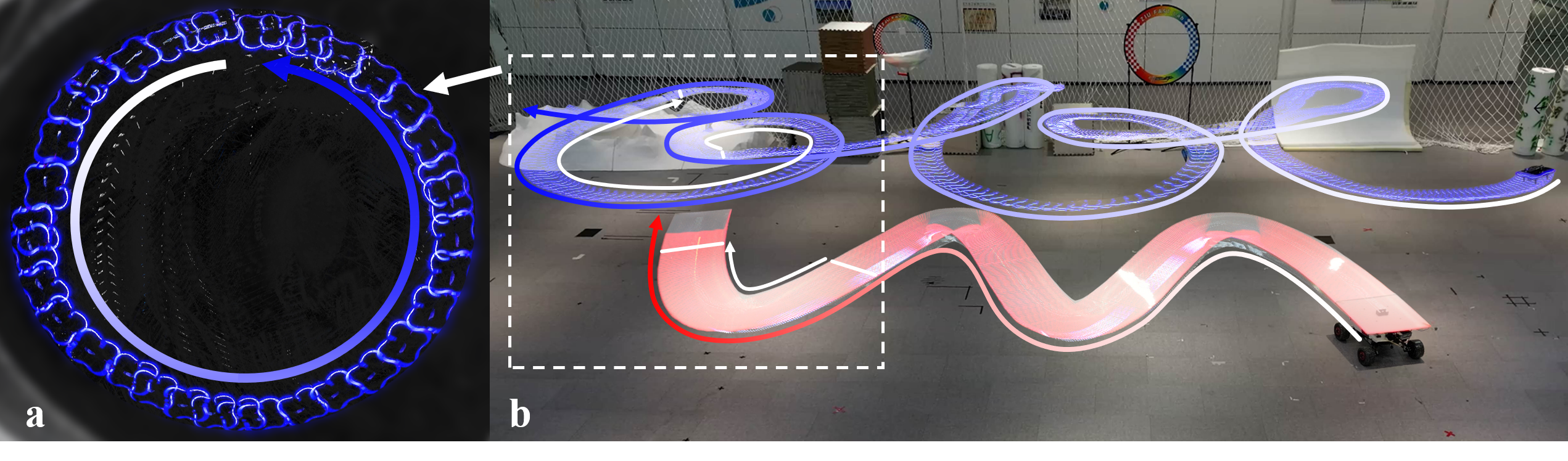}
\captionof{figure}{A quadrotor orbits a UGV by applying CoNi-MPC controller with a pre-computed circular trajectory in the UGV non-inertial frame. 
        (a) is the accumulated shots of the quadrotor from the view of a camera on the UGV, which shows the relative circular trajectory of the quadrotor.
        (b) shows the experiment from a third-person view in the world frame, in which the flight trajectory appears chaotic along with the UGV S-shape trajectory.}
\label{top_figure}
\end{center} 
}]

\begingroup
\renewcommand\thefootnote{\IEEEauthorrefmark{2} }
\footnotetext{\textbf{Equal contribution}}
\renewcommand\thefootnote{}
\footnotetext{This work was supported by National Nature Science Foundation of China under Grant 62103368. (Corresponding author: Yanjun Cao, Chao Xu.\tt\footnotesize \{yanjunhi, cxu\}@zju.edu.cn)}
\renewcommand\thefootnote{\textsuperscript{1} }
\footnotetext{Huzhou Institute of Zhejiang University, Huzhou, 313000, China.}
\renewcommand\thefootnote{\textsuperscript{2} }
\footnotetext{State Key Laboratory of Industrial Control Technology, Institute of Cyber-Systems and Control, Zhejiang University, Hangzhou, 310027, China.}
\renewcommand\thefootnote{\textsuperscript{3} }
\footnotetext{The Chinese University of Hong Kong, Shenzhen, 518172, China.}
\renewcommand\thefootnote{\textsuperscript{4} }
\footnotetext{Department of Computer Engineering and Software Engineering, Polytechnique Montreal, Canada.}
\endgroup

\begin{abstract}

  This paper presents a novel solution for UAV control in cooperative
  multi-robot systems, which can be used in various scenarios such as
  leader-following, landing on a moving base, or specific relative motion with a
  target. Unlike classical methods that tackle UAV control in the world frame,
  we directly control the UAV in the target coordinate frame, without making
  motion assumptions about the target. In detail, we formulate a non-linear
  model predictive controller of a UAV, referred to as the agent, within a
  non-inertial frame (i.e., the target frame). The system requires the relative
  states (pose and velocity), the angular velocity and the accelerations of the
  target, which can be obtained by relative localization methods and ubiquitous
  MEMS IMU sensors, respectively. This framework eliminates dependencies that
  are vital in classical solutions, such as accurate state estimation for both
  the agent and target, prior knowledge of the target motion model, and
  continuous trajectory re-planning for some complex tasks. We have performed
  extensive simulations to investigate the control performance with
  varying motion characteristics of the target. Furthermore, we conducted
  real robot experiments,
{ 
  employing either simulated relative pose estimation from motion capture
  systems indoors or directly from our previous relative pose estimation devices
  outdoors}, to validate the applicability and feasibility of the proposed
approach.

\end{abstract}


\vspace{0.60cm}
\section{Introduction}

\IEEEPARstart{R}{ecently}, quadrotors or drones, due to their agility and
lightweight nature, have been widely used in surveillance, search-and-rescue,
and cinematography. The rapid development has led to a growing demand for
multi-robot systems such as UAV-UGV pairs~\cite{uav_ugv}, leader-follower
systems~\cite{leader_follower}, multi-agent formation~\cite{quan2022formation},
autonomous landing~\cite{intro_landing1,intro_landing2}, etc. This paper focuses
on an air-ground robot system in which the UAV (referred to as the
agent/quadrotor/drone/follower) is actively controlled to fulfill a task along
with an independently controlled UGV (referred to as the target/base/leader).
State estimation, planning, and controllers are crucial components in developing
versatile systems for interactive or cooperative tasks. In classical pipelines,
relative state, typically obtained through direct mutual measurements or
subtraction from global state estimations, is used as a feedback to control the
motion of UAVs in the world frame \cite{niu_vision_landing}. In these pipelines,
controllers require a complete system model of the quadrotor. Furthermore, in
complex tasks such as in \cite{fast_tracker,perching_one}, appropriate
trajectories and continuous re-planning are needed to achieve good performance.

To achieve good performance for the air-ground system, current
state-of-the-art air-ground (agent-target) collaborative planning-control
systems such as \cite{fast_tracker,perching_one} have to face the following challenges:
\begin{itemize}
\item Accurate \textbf{absolute state estimations} for both the agent and the
target to achieve demanding high-precision relative motion planning and
control, which is hard to be guaranteed in challenging environments (GPS
denied, feature-less) or for long-term tasks (accumulated drifts);
\item A \textbf{prior kinematic model} of the target is necessary to be known
by the agent to predict the target's movements, which may fail if the given
model is not accurate or the assumptions of the target model do not hold;
\item  \textbf{Continuous trajectory re-planning} of the agent is needed to be responsive and adaptive to the target's motions, which can lead to heavy computation loads.
\end{itemize} 
Therefore, we propose CoNi-MPC that directly controls the agent in a target's
body frame utilizing relative estimations and target's IMU data, eliminating all
dependencies in absolute world frame.

Typically, a full SLAM stack that fuses multiple sensors (vision, lidar,
GPS, IMU, etc.) is used to acquire accurate state estimation. However, SLAM
algorithms generally demand high computational cost and rely on good environment
features to achieve robust estimation. Maintaining long-term SLAM for a system
that only requires interactive actions may be also considered redundant. At the
same time, having prior knowledge of the kinematic model of the moving target is
necessary for the agent to predict the target's following trajectory accurately.
However, this could be difficult to be guaranteed considering the target's
individual tasks and motion. Even with accurate state estimation and a precise
kinematic model, the dynamic evolution of both the target's state and the
agent's state demands continuous trajectory re-planning.

To overcome these challenges {and directly control the agent in the
  target's body frame}, we design CoNi-MPC, a novel systematic solution that
formulates a non-linear model predictive controller of a UAV within a
non-inertial frame, specifically the target frame. The system only requires the
relative states (pose and velocity) , the angular velocity and the accelerations
of the target, which can be obtained by relative localization methods and
ubiquitous MEMS IMU sensors, respectively. This solution eliminates the
dependency on state estimation information in the world frame{, and
  only requires relative estimation}. We directly controls the UAV in the target
coordinate frame without making any motion assumptions about the target.
Additionally, the system does not require trajectory re-planning even for some
complex tasks.

This CoNi-MPC framework can be directly applied to various application tasks,
such as leader-following, directional landing, and complex relative motion
control. All these tasks can be implemented by changing the
reference within the model. In the leader-follower control, a single fixed
relative point within the leader's frame serves as an input to guide the
follower. For landing or more complex inter-robot interactive tasks, the agent
control only requires one pre-computed trajectory relative to the target,
without any re-planning requirement. Fig.~\ref{top_figure} shows snapshots
of a drone circling over a ground vehicle (orbit flight), where the drone is
controlled in the ground vehicle's frame (non-inertial frame) and the vehicle follows
an S-shape trajectory (unknown to the drone). The drone's trajectory traces a
circle, as viewed from the vehicle's perspective, while the trajectory of the
drone in the world frame is rather complex.



{To the best of our knowledge, this is the first work realizing complex
  interactions between an agent and a target that only requires relative position
  estimation and the target's IMU data. The contributions of our work are:
\begin{itemize}
\item We propose a systematic framework for drone-target relative motion control
  using MPC by fully modeling the drone in the target's non-inertial body frame.
  The system does not need to know the absolute pose and motion of target in the
  world frame.
\item With the relative motion model, we group the target-dependent elements
  together and substitute it with IMU information from target body frame. This
  operation eliminates the dependency on the data in the world frame and makes
  this method feasible in real-world setup.
\item The proposed MPC controller works as a unified framework supporting
  various UAV-target interaction tasks (eg. leader-following, aggressive
  directional landing, dynamic rings crossing, and orbit flight) with high
  tracking accuracy while eliminating continuous trajectory re-planning.
\end{itemize}}

\vspace{-0.20cm}
\section{Related Work}

Autonomous landing \cite{niu_vision_landing, wang_vision_landing},
leader-following systems \cite{LeaderFollower01,LeaderFollower02}, and tracking
\cite{Falanga_pampc} have been extensively investigated individually,
considering their unique task characteristics and challenges. Niu et al.
\cite{niu_vision_landing} introduce a vision-based autonomous landing method for
UAV-UGV cooperative systems. They employ multiple QR codes on the landing pad of
the UGV to obtain estimations of relative distance, velocity and direction
between the two vehicles, as well as UAV state from Visual Inertial Odometry
(VIO) or GPS. Based on these estimations, a velocity controller utilizing a
control barrier function (CBF) and a control Lyapunov function (CLF) are
designed for the quadrotor landing on the moving UGV. Wang et al.
\cite{wang_vision_landing} propose a systematic approach for vision-based
autonomous landing that utilizes EKF and VIO for pose estimation and a similar
marker detection method obtaining the relative information between the UAV and
the UGV. Han et al. \cite{LeaderFollower01} utilize a complex Laplacian based
similar formation control algorithm over leader-follower networks with the
actively estimated relative position. Giribet et al. \cite{LeaderFollower02}
propose a tracking controller based on dual quaternion pose representations and
cluster-space in a leader-follower task, with the objective of minimizing
steady-state error. The UAVs in these works are all controlled in the world
frame and therefore the global state estimation is essential for the system. In
our system, we formulate the model in a pure relative motion control in the
target frame. The system avoids involving the state estimation in the world
frame, which can be difficult or fragile under specific conditions.

While relative motion based control is more widely used in the field of space
technology, such as target tracking and docking for approaching operations
\cite{spacecraft}, a limited number of works in robotics explore modeling and
controlling relative motion for UAV-UGV cooperation, particularly in the context
of quadrotor control. Marani et al.~\cite{relative_easy} investigate the
dynamics of a quadrotor in a non-inertial frame without rotation,
assuming the referencing non-inertial frame only performs translational movement,
and they use a sliding mode controller for trajectory tracking. Jin et al.
\cite{relative_hard} investigate the relative motion model of a quadrotor in a
non-inertial frame and propose two controllers (relative position and attitude
controllers) for a quadrotor landing on a moving vessel. Although they directly
address relative motion constraints, the control input remains reliant on
attitude in a world frame, necessitating global state estimation. Li et al.
\cite{visual_servoing_robocentric} propose a robocentric model-based visual
servoing method for hovering and obstacle avoidance for a single drone,
employing model predictive control. Their method constructs the ``relative''
states in the drone's body frame by using the RGB-D camera to detect
targets, which eliminates the state dependency on the world frame.
{DeVries et al. \cite{noninertial_formation} proposed a distributed
  formation controller in a non-inertial reference frames but still map the
  agent's states and control inputs to an inertial frame.}

The model predictive control framework serves as the base of many works related
to direct control and trajectory planning in the literature. Falanga et
al.~\cite{Falanga_pampc} propose a non-linear MPC (PAMPC) method for quadrotors,
combining perception and action terms into the optimization. A VIO estimator and
the PAMPC method are applied to allow the quadrotor to follow a trajectory while
maintaining a point of interest in its field of view. Ji et al.~\cite{CMPCC}
propose a disturbance-adaptive receding horizon low-level replanner for
autonomous drones, which can generate collision-free and temporally
optimized local reference trajectories. Similarly, Romero et al.~\cite{MPCC} handle the
problem of generating temporal optimized trajectories for quadrotors. The
proposed MPCC also integrates temporal optimization into the standard MPC
formulation to solve the time allocation problem online. {In
  \cite{SNMPC}, a stochastic and predictive MPC (SNMPC) is proposed to minimize
  the total amount of uncertainty in the target observation and the robot
  state estimation, to effectively maintain the desired pose of the robot
  relative to the moving target.}

The aforementioned works or their applications in multi-robot cooperation still
require global state estimation and frequent global path re-planning. Our work,
based on relative estimation, is inherently suitable for cooperation
tasks without requiring global state estimation. Moreover, costly global
path re-planning can be avoided thanks to the system model in the non-inertial
frame.

\vspace{-0.20cm}
\section{Problem Formulation and CoNi-MPC}

We consider a cooperative system consisting of a UGV as the target and a UAV as the agent. The objective is to regulate the motion of the UAV in conjunction with the UGV for multi-robot cooperation tasks, such as leader-follower, landing, orbit flight, etc. 

\vspace{-0.25cm}
\subsection{Notations}

\begin{figure}[!t]
\setlength{\belowcaptionskip}{-0.25cm}
\centering
\includegraphics[width=.5\textwidth]{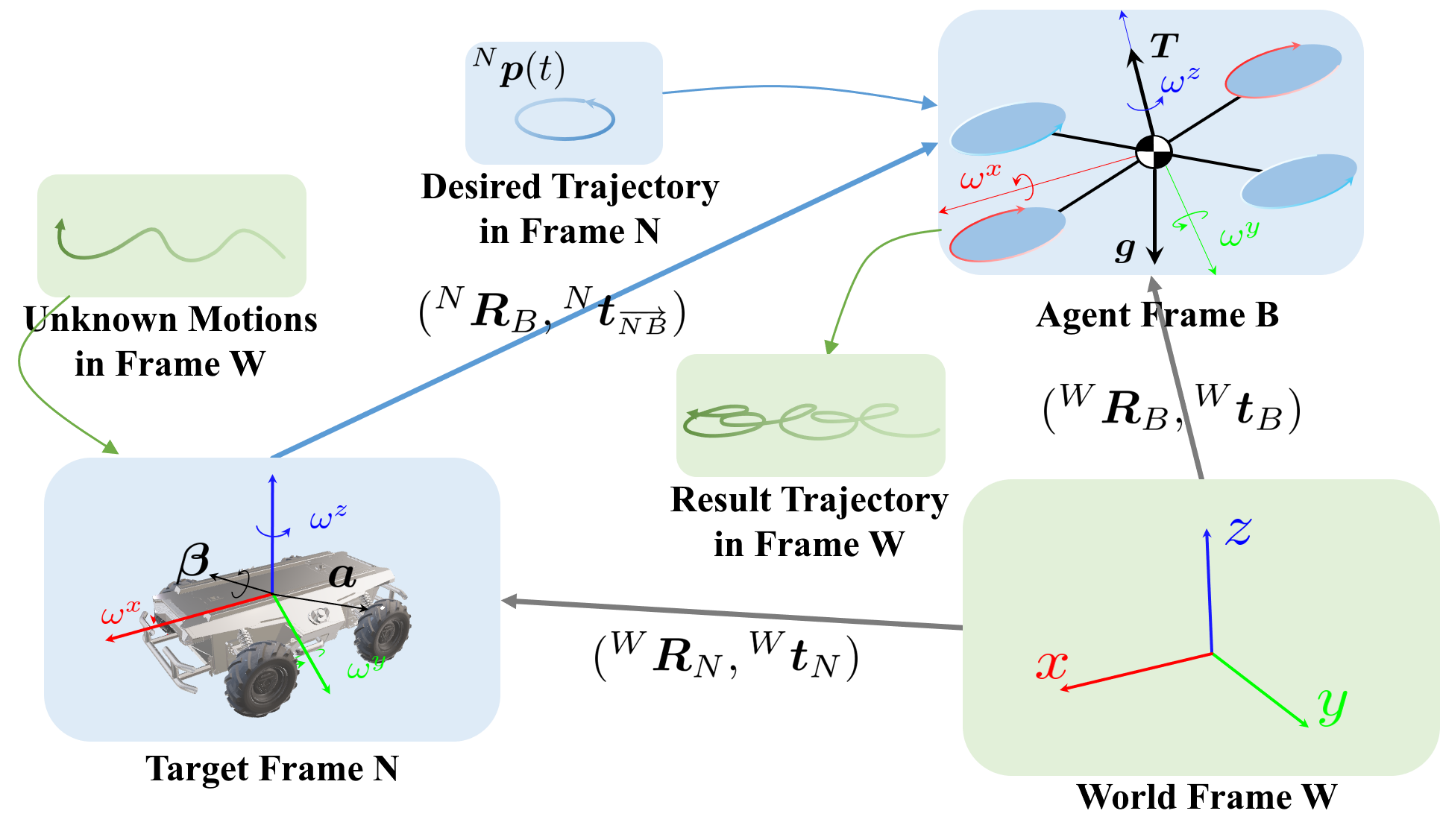}
\caption{The transformation relationship among the quadrotor (the agent) body frame, the non-inertial (the target) frame, and the world frame. The target is controlled externally, and the agent is controlled by feeding reference (e.g. trajectory) defined in the target's frame.}
\label{relative}
\end{figure}

As shown in Fig. \ref{relative}, we define the agent frame as $B$ attached to the body frame of the quadrotor, the target frame as $N$ attached to the body frame of the UGV, and frame $W$ is the inertial world frame. 
We denote scalar numbers with lowercase letters, vectors with bold lowercase letters, and matrices with bold uppercase letters. 
The left superscript indicates the coordinate system where the variable is expressed.  
If no other specification, values without any left superscript are expressed in the world frame $W$. 
For example, we denote the relative position of frame $B$ w.r.t. frame $N$ (non-inertial) by ${}^N\boldsymbol{p}_B$, the relative velocity by ${}^N\boldsymbol{v}_B$, and the relative orientation by ${}^N\boldsymbol{q}_B$. 
The right superscript $x$, $y$, or $z$ on a vector means the element of the vector, e.g., $\boldsymbol{t}^x$ is the $x$ element of $\boldsymbol{t}$. 
$\odot$ means quaternion Hamilton product. 
The skew-symmetric matrix of a vector $\boldsymbol{t}$ is denoted as $[\boldsymbol{t}]_{\times}$.
The measured vector $\boldsymbol{v}$ from sensors is denoted as $\widehat{\boldsymbol{v}}$.
Table \ref{notations} lists the main notations used in this paper.
\begin{table}[t]
\caption{Table of Notations}
\begin{center}
\label{notations}
\begin{tabular}{r c p{.65\linewidth}}
\toprule
${}^N\boldsymbol{p}_B$ & $\triangleq$ & Relative position of the agent in the target's frame \\
${}^N\boldsymbol{v}_B$ & $\triangleq$ & Relative velocity of the agent in the target's frame \\
${}^{*}\boldsymbol{q}_{\#}$ & $\triangleq$ & Unit quaternion from $\#$ to $*$ \\
${}^{*}\boldsymbol{R}_{\#}$ & $\triangleq$ & Rotation matrix from $\#$ to $*$\\
${}^{*}\boldsymbol{t}_{\sss \uvec{NB}}$ & $\triangleq$ & Translation vector from $N$ to $B$ expressed in $*$\\
$\boldsymbol{t}_{\#}$ & $\triangleq$ & Translation vector from $W$ to $\#$ expressed in $W$\\
$\boldsymbol{g}$ & $\triangleq$ & Gravitational acceleration\\
${}^B\boldsymbol{T}_B$ & $\triangleq$ & Normalized collective thrust of $B$, system input\\
${}^B\boldsymbol{\Omega}_B$ & $\triangleq$ & Body rate of $B$, system input\\
${}^N\boldsymbol{a}_N$ & $\triangleq$ & Linear acceleration of $N$ expressed in $N$ \\
${}^N\boldsymbol{\Omega}_N$ & $\triangleq$ & Body rate (angular velocity) of $N$\\
${}^N\boldsymbol{\beta}_N$ & $\triangleq$ & Angular acceleration of $N$\\
$(r, v, \omega)$ & $\triangleq$ &  Experiment parameter configuration\\
\bottomrule
\end{tabular}
\end{center}
\vspace{-0.50cm}
\end{table}

\vspace{-0.25cm}
\subsection{Quadrotor System Model in Non-Inertial Frame}

{In this section, we derive the quadrotor system model in the non-inertial frame $N$ by introducing an intermediate inertial world frame $W$. Our derivation shows that all the dependencies on this world frame are eliminated at the end.
Fig. \ref{relative} shows the relationship among the three frames. 
The relative position is:}
\begin{equation}
{}^N\boldsymbol{p}_B = {}^N\boldsymbol{R}_W \boldsymbol{t}_{\sss \uvec{NB}} 
\end{equation}
where $\boldsymbol{t}_{\sss \uvec{NB}} = \boldsymbol{t}_B - \boldsymbol{t}_N$ is the translational vector pointing from the origin of frame $N$ to frame $B$.
Then we get the relative velocity by applying a time derivative as following
\begin{equation}
\begin{aligned}
\label{pdot}
{}^N\dot{\boldsymbol{p}}_B = {}^N\boldsymbol{v}_B
&= \frac{d}{dt}({}^N\boldsymbol{R}_{W}) \boldsymbol{t}_{\sss \uvec{NB}} + {}^N\boldsymbol{R}_{W} \dot{\boldsymbol{t}}_{\sss \uvec{NB}} \\
&=-[{}^N\boldsymbol{\Omega}_N]_{\times} {}^N\boldsymbol{R}_{W} \boldsymbol{t}_{\sss \uvec{NB}} + {}^N\boldsymbol{R}_{W} \dot{\boldsymbol{t}}_{\sss \uvec{NB}}\\
&=-[{}^N\boldsymbol{\Omega}_N]_{\times} {}^N\boldsymbol{p}_B + {}^N\boldsymbol{R}_{W} \dot{\boldsymbol{t}}_{\sss \uvec{NB}}\\
\end{aligned}
\end{equation}
The relative acceleration is the time derivative of the relative velocity
\begin{equation}
\begin{aligned}
\label{vdot}
{}^N\dot{\boldsymbol{v}}_B
&= -\frac{d}{dt}([{}^N\boldsymbol{\Omega}_N]_{\times}) {}^N\boldsymbol{p}_B - [{}^N\boldsymbol{\Omega}_N]_{\times} {}^N\dot{\boldsymbol{p}}_B \\
&- [{}^N\boldsymbol{\Omega}_N]_{\times}{}^N\boldsymbol{R}_{W} \dot{\boldsymbol{t}}_{\sss \uvec{NB}} + {}^N\boldsymbol{R}_{W} \ddot{\boldsymbol{t}}_{\sss \uvec{NB}}\\
&= -\frac{d}{dt}([{}^N\boldsymbol{\Omega}_N]_{\times}){}^N\boldsymbol{p}_B - [{}^N\boldsymbol{\Omega}_N]_{\times}{}^N\boldsymbol{v}_B\\
&- [{}^N\boldsymbol{\Omega}_N]_{\times}({}^N\boldsymbol{v}_B +[{}^N\boldsymbol{\Omega}_N]_{\times} {}^N\boldsymbol{p}_B) + {}^N\boldsymbol{R}_{W} (\ddot{\boldsymbol{t}}_{B} - \ddot{\boldsymbol{t}}_{N})\\
&= -[{}^N\boldsymbol{\beta}_N]_{\times}{}^N\boldsymbol{p}_B - 2[{}^N\boldsymbol{\Omega}_N]_{\times}{}^N\boldsymbol{v}_B - [{}^N\boldsymbol{\Omega}_N]_{\times}^2 {}^N\boldsymbol{p}_B \\
&+ {}^N\boldsymbol{R}_{B}{}^B\boldsymbol{T}_B + \underbrace{{}^N\boldsymbol{R}_{W}\boldsymbol{g}  - {}^N\boldsymbol{R}_{W}\boldsymbol{a}_N}_{\text{values relying on estimations in } W}
\end{aligned}
\end{equation}
where $\ddot{\boldsymbol{t}}_{B}$ in Equ. \ref{vdot} is the acceleration of a quadrotor modeled in the world frame as a rigid body as in \cite{quadrotor_model} 
\begin{equation}
\ddot{\boldsymbol{t}}_{B} = {}^W\boldsymbol{R}_B {}^B\boldsymbol{T}_B + \boldsymbol{g}
\end{equation}
${}^B\boldsymbol{T}_B = [0, 0, T]^\top$ is the normalized collective thrust of the quadrotor and $T = \sum_i T_i, i\in \{1,2,3,4\}$ is the normalized thrust force from four motors, 
$\boldsymbol{g} = [0, 0, -g]^\top$ is the gravity, 
${}^N\boldsymbol{\Omega}_N$ is the body rate of the non-inertial frame, 
${}^N\boldsymbol{\beta}_N$ is the angular acceleration of the non-inertial frame, 
${}^N\boldsymbol{R}_W\boldsymbol{a}_N$ is the linear acceleration of the non-inertial frame expressed in the non-inertial frame. 

The rotation matrix from $B$ to $N$ is 
\begin{equation}
{}^N\boldsymbol{R}_{B} = {}^N\boldsymbol{R}_{W} {}^W\boldsymbol{R}_{B}
\end{equation}
The time derivative of the above rotation matrix is 
\begin{equation}
\begin{aligned}
{}^N\dot{\boldsymbol{R}}_{B}
&= \frac{d}{dt}({}^N\boldsymbol{R}_{W}){}^W\boldsymbol{R}_{B} + {}^N\boldsymbol{R}_{W} \frac{d}{dt}({}^W\boldsymbol{R}_{B}) \\
&= -[{}^N\boldsymbol{\Omega}_N]_{\times} {}^N\boldsymbol{R}_{B} + {}^N\boldsymbol{R}_{B}[{}^B\boldsymbol{\Omega}_B]_{\times} 
\end{aligned}
\end{equation}
At the same time, we show the quaternion here for model implementation in the next section 
\begin{equation}
\label{rotdot}
\begin{aligned}
{}^N\dot{\boldsymbol{q}}_B
&= {}^N\dot{\boldsymbol{q}}_W \odot {}^W\boldsymbol{q}_B + {}^N\boldsymbol{q}_W \odot {}^W\dot{\boldsymbol{q}}_B \\
&= -\frac{1}{2}{}^N\boldsymbol{\Omega}_N \odot {}^N\boldsymbol{q}_{B} + \frac{1}{2} {}^N\boldsymbol{q}_{B} \odot {}^B\boldsymbol{\Omega}_B
\end{aligned}
\end{equation}

In the system model of Equ. \ref{vdot} and Equ. \ref{rotdot}, we almost eliminate the dependency on values in world frame $W$ except for $^N\boldsymbol{R}_{W}$ in Equ. \ref{vdot}.
We notice that the last two terms (${}^N\boldsymbol{R}_{W}\boldsymbol{g}  - {}^N\boldsymbol{R}_{W}\boldsymbol{a}_N$) in Equ. \ref{vdot} are actually the total measured acceleration of target expressed in target's frame, which 
can be handled using the data from an IMU attached to the non-inertial frame to the relative system model directly.  
In detail, Equ. \ref{vdot} contains the projected gravitational acceleration ${}^N\boldsymbol{R}_W \boldsymbol{g}$ and the acceleration of the non-inertial frame ${}^N\boldsymbol{a}_N$($= {}^N\boldsymbol{R}_{W}\boldsymbol{a}_N$). 
The true acceleration of a MEMS IMU sensor can be calculated by applying the negative gravity in the target's body frame as 
\begin{equation}
\label{eq:IMU}
{}^{N} \boldsymbol{a}_N =
{}^{N} \boldsymbol{R}_W
\prescript{W}{}{
\begin{bmatrix}
0\\
0\\
-g
\end{bmatrix}
}
+
\prescript{N}{}{
\begin{bmatrix}
\hat a^x\\
\hat a^y\\
\hat a^z
\end{bmatrix}
}
\end{equation}
where $\hat a^x$, $\hat a^y$, and $\hat a^z$ are the measured acceleration data from the IMU. 
With Equ. \ref{eq:IMU}, Equ. \ref{vdot} can be reformulated to 
\begin{equation}
\label{eq:v_dot_imu}
\begin{aligned}
{}^N\dot{\boldsymbol{v}}_B
&= -[{}^N\boldsymbol{\beta}_N]_{\times} {}^N\boldsymbol{p}_B - 2 [{}^N\widehat{\boldsymbol{\Omega}}_N]_{\times} {}^N\boldsymbol{v}_B - [{}^N\widehat{\boldsymbol{\Omega}}_N]_{\times}^2 {}^N\boldsymbol{p}_B \\
&+ {}^N\boldsymbol{R}_{B}{}^B\boldsymbol{T}_B - {}^N\widehat{\boldsymbol{a}}_{N}
\end{aligned}
\end{equation}
where ${}^N\widehat{\boldsymbol{a}}_{N}$ and ${}^N\widehat{\boldsymbol{\Omega}}_{N}$ are the measured linear acceleration and angular velocity from the IMU, respectively. 

\vspace{-0.25cm}
\subsection{CoNi-MPC}

We propose a \textbf{Co}operative \textbf{N}on-\textbf{i}nertial Frame Based \textbf{M}odel \textbf{P}redictive \textbf{C}ontrol (CoNi-MPC) with the above system model targeting relative motion control.
We define the cooperative system state 
$ \boldsymbol{x} = [ {}^N\boldsymbol{p}_B;\; {}^N\boldsymbol{v}_B;\; {}^N\boldsymbol{q}_B;\; {}^N\widehat{\boldsymbol{a}}_N;\; {}^N\widehat{\boldsymbol{\Omega}}_N;\; {}^N\boldsymbol{\beta}_N ] \in \mathbb{R}^{19} $.
The first three vectors ${}^N\boldsymbol{p}_B$, ${}^N\boldsymbol{v}_B$, and ${}^N\boldsymbol{q}_B$, as defined in above section, are the relative quantities in the system.
The last three vectors ${}^N\widehat{\boldsymbol{a}}_N$, ${}^N\widehat{\boldsymbol{\Omega}}_N$, and ${}^N\boldsymbol{\beta}_N$ contain the dynamic information of the non-inertial frame. 
The time derivative of the angular velocity is $\frac{d}{dt}({}^N\widehat{\boldsymbol{\Omega}}_N)={}^N\boldsymbol{\beta}_N$. 
For the linear and angular accelerations, we assume their change rates are 0 in each control window, i.e., ${}^N\dot{\widehat{\boldsymbol{a}}}_N=0$ and $\dot{\boldsymbol{\beta}}_N=0$. 
As they are relatively high order values, this assumption does not affect the performance very much for our system.   
We put the dynamic information of the non-inertial frame in the state vector for convenience in the implementation part and for future improvement on a actively collaborative system (expanding the control vector and adding the dynamic evolution of frame $N$). 
The control input is
$ \boldsymbol{u} = [ T;\; {}^B\Omega_{B}^x;\; {}^B\Omega_{B}^y;\; {}^B\Omega_{B}^z ] \in \mathbb{R}^{4} $ where $T = \sum_{i=1}^4T_i$.

We define the quadratic cost
\begin{equation*}
\boldsymbol{C}(\boldsymbol{x}, \boldsymbol{u}) = \lVert \boldsymbol{x}(t) - \boldsymbol{x}(t)_{ref} \rVert_{\boldsymbol{Q}} + \lVert \boldsymbol{u}(t) - \boldsymbol{u}_h \rVert_{\boldsymbol{R}}
\end{equation*}
where $\lVert \boldsymbol{x}\rVert_{\boldsymbol{M}} = \boldsymbol{x}^\top \boldsymbol{M} \boldsymbol{x}$ and $\boldsymbol{u}_h = [g;0;0;0]$ is the hover input. 
The discretized optimization problem is 
\begin{equation}
\begin{aligned}
\min_{\boldsymbol{u}_0, \dots, \boldsymbol{u}_{N-1}} \quad & \sum_{k=0}^{N-1} (
\lVert \boldsymbol{x}(k) - \boldsymbol{x}(k)_{ref} \rVert_{\boldsymbol{Q}} +
\lVert \boldsymbol{u}(k) - \boldsymbol{u}_h \rVert_{\boldsymbol{R}} ) \\  
&+ \lVert \boldsymbol{x}(N) - \boldsymbol{x}(N)_{ref} \rVert_{\boldsymbol{Q}_{final}}\\
\textrm{s.t.} \quad & \boldsymbol{x}(0) = \boldsymbol{x}_0 \\
&\boldsymbol{x}(k+1) = f_d(\boldsymbol{x}(k), \boldsymbol{u}(k)) \\
& T_{\text{min}} \leq T \leq T_{\text{max}} \\
& \lVert {}^B\Omega_B^x \rVert \leq \Omega_{\text{rp}} \\
& \lVert {}^B\Omega_B^y \rVert \leq \Omega_{\text{rp}} \\
& \lVert {}^B\Omega_B^z \rVert \leq \Omega_{\text{yaw}} \\
\end{aligned}
\end{equation}
where $\boldsymbol{x}_0$ is the state estimation of the system in each control iteration, $f_d$ is the discretized system model, $T_{\text{min}}$ is the minimum thrust, $T_{\text{max}}$ is the maximum thrust constraint, $\Omega_{\text{rp}}$ is the maximum roll and pitch angular speed, and $\Omega_{\text{yaw}}$ is the maximum yaw angular speed. 

As our system does not rely on any information in the global world frame and only relates to the agent and target, the system can handle the cooperation between robots elegantly.
Tasks like leader-follower, landing, orbit flight, rings crossing, etc. can be solved by simply defining the reference in the CoNi-MPC.
The advantage of the system is that the reference is fixed, which is just a pre-computed expression of the relative motion between robots and does not need any online replaning. 
We classify the reference into two categories, fixed-point scheme and fixed-plan scheme, corresponding to leader-follower and complex motion respectively.

\subsubsection{Fixed point scheme (leader and follower)}

The proposed method can be easily used for leader-follower control.
CoNi-MPC only needs a fixed point (containing the full state) so as to let the agent track that point while the target moves. 
For example, an array containing the same point reference, $\boldsymbol{x}(k) = [(0, 0, z), \boldsymbol{0}^\top, (1, 0, 0, 0), \boldsymbol{0}^\top, \boldsymbol{0}^\top, \boldsymbol{0}^\top]^\top$, fed to the controller will let the agent hover at the ${}^N\boldsymbol{p}_B=(0, 0, z)^\top$ point in the non-inertial target frame with the same orientation of the target frame, even if the target frame may arbitrarily move in the world frame.

\subsubsection{Fixed plan scheme (complex trajectories)}

For complex tasks such as landing, orbit flight, and rings crossing, we simply need to define the trajectory of the agent in the target frame. 
The landing tasks can use a trajectory approaching the origin of $N$ frame and the orbit flight is a circle trajectory directly. 
We adopt our previous work of a minimum control effort polynomial trajectory class named MINCO \cite{MINCO} to define the relative trajectory between robot.
MINCO trajectory can achieve smooth motions by decoupling the space and time parameters of the trajectory for users, which greatly improves the quality and efficiency of trajectory generation.
We only need to take initial and terminal relative states of the UAV as boundary conditions, 
and specify the position of intermediate waypoints and the time duration of each piece to obtain a polynomial trajectory $\boldsymbol{p}({t})$ with minimum jerk.
Furthermore, we limit the maximum relative velocity and acceleration to guarantee dynamic feasibility.
After discretizing $\boldsymbol{p}({t})$ and calculating the orientation based on the differential flatness of multicopters \cite{Kumar_flat}, we can get a series of reference states $\{[{}^N\boldsymbol{p}_B(k)\;;{}^N\boldsymbol{v}_B(k)\;;{}^N\boldsymbol{q}_B(k)]\}_{k=0}^{N}$.
Thus, the proposed controller can be fed with only one fixed global trajectory to achieve autonomous landing and tracking while the UGV moves.

\vspace{-0.50cm}
\subsection{Implementation}

\begin{figure}[!t]
\centering
\setlength{\belowcaptionskip}{-0.50cm}
\includegraphics[width=0.5\textwidth]{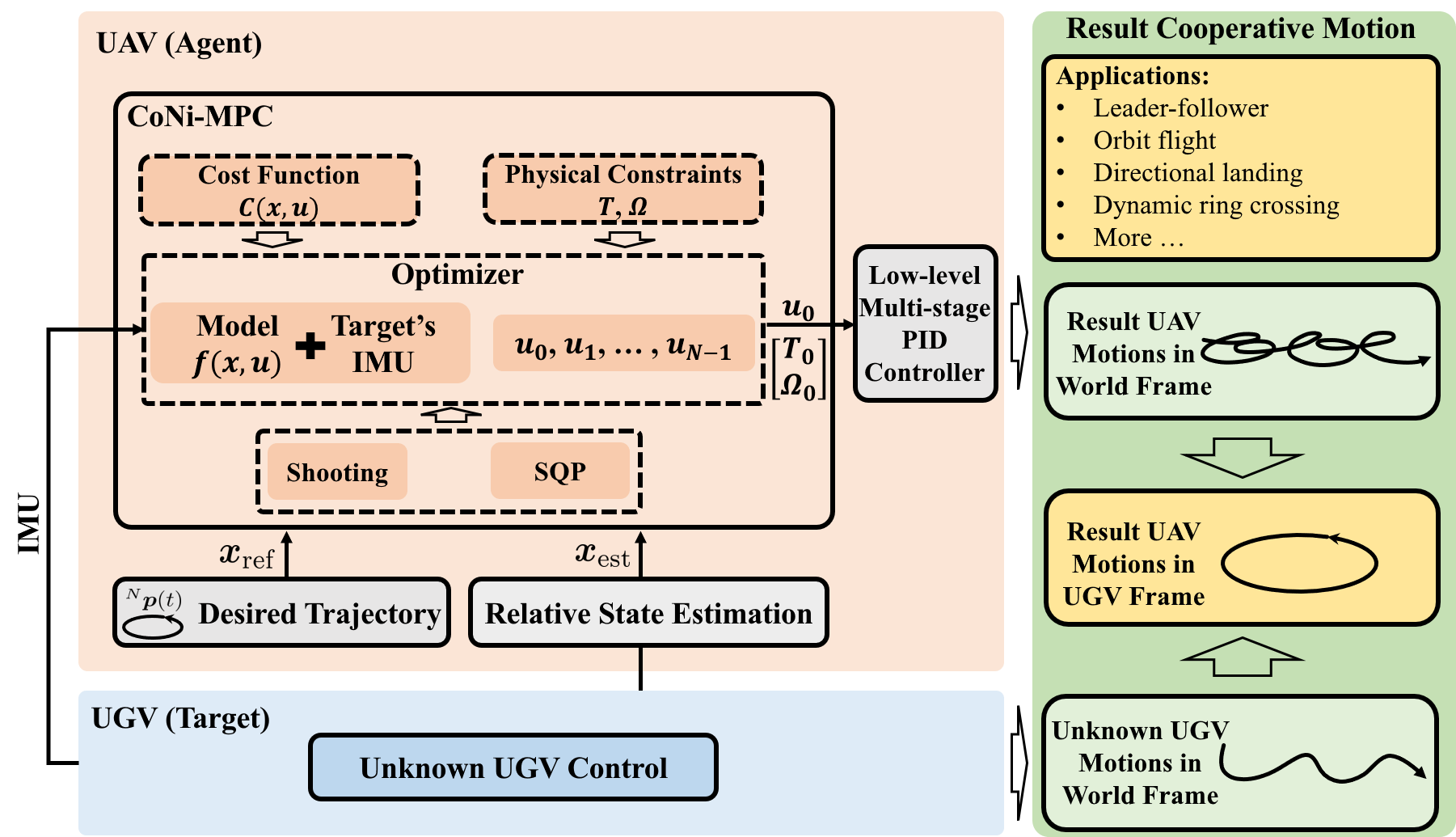}
\caption{System overview and implementation of UAV-UGV cooperative motion control using CoNi-MPC.}
\label{pipeline}
\end{figure}

Fig. \ref{pipeline} shows the system overview of UAV-UGV cooperative motion control system.
A CoNi-MPC controller, implemented using ACADO toolkit \cite{ACADO}, is employed by the UAV and works as a high-level controller to produce thrust and body rate control command $\boldsymbol{u}_0$.
A low-level multi-stage PID controllers is used to track the control command. 
CoNi-MPC needs a pre-defined desired relative motion trajectory refer to the UGV as motion control target.
With input of UGV's IMU measurements and relative state estimations, CoNi-MPC solves a optimization via multiple shooting technique and Runge-Kutta integration scheme.
An average signal filter is applied to the IMU data from the non-inertial frame, {which is transmitted through ROS's multi-machine communication mechanism in current setup.}  
The relative estimation can be generated either from a motion capture system or directly from our previous work of CREPES \cite{CREPES}, a relative estimation device. 
For each control iteration of the MPC optimization problem, we set the time window as $T = 2$ seconds and discretization time step $dt = 0.1$ second.
{The real control loop time of the MPC is around 10 ms, which is smaller than $dt=100$ ms. 
This implementation differs from standard MPC formulation, where CoNi-MPC uses latest relative state to produce the control command much more frequently, but with relatively small scale optimization problem to save computation cost.}
In each iteration, the initial state is set as the current estimated relative state $\boldsymbol{x}_{\text{est}}$.
For ${}^N\widehat{\boldsymbol{a}}_N$, ${}^N\widehat{\boldsymbol{\Omega}}_N$, and ${}^N\boldsymbol{\beta}_N$, the penalty terms for them are set to $\boldsymbol{Q}(i,i)=0, \;\forall i = 10\dots19$. 
Note that in the simulation and experiment, since the angular acceleration is hard to retrieve, we set this term to $\boldsymbol{0}$ both in estimations and references, which assumes that the non-inertial frame rotates with a constant angular velocity in each prediction horizon.

\vspace{-0.25cm}
\section{Experiments}

\begin{figure*}[!t]
\centering
\setlength{\belowcaptionskip}{-0.25cm}
\includegraphics[width=\textwidth]{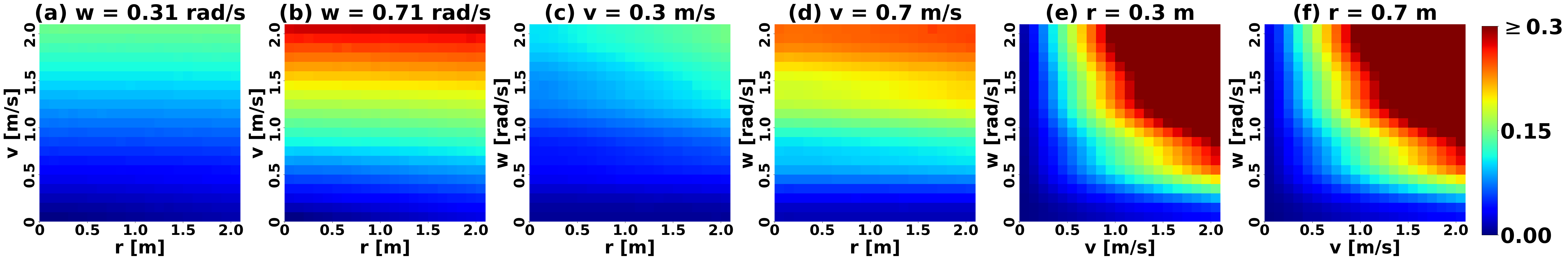}
\caption{Mean errors for tracking a point with different $(r,v,\omega)$ settings.
        (a-b) error distribution of $r-v$ with $\omega=$ 0.31 and 0.71. (c-d) error distribution of $\omega-r$ with $v=$ 0.3 and 0.7. (e-f) error distribution of $\omega-v$ with $r=$ 0.3 and 0.7.}
\label{fixed_point}
\end{figure*}
\begin{figure*}[!t]
\centering
\setlength{\belowcaptionskip}{-0.50cm}
\includegraphics[width=\textwidth]{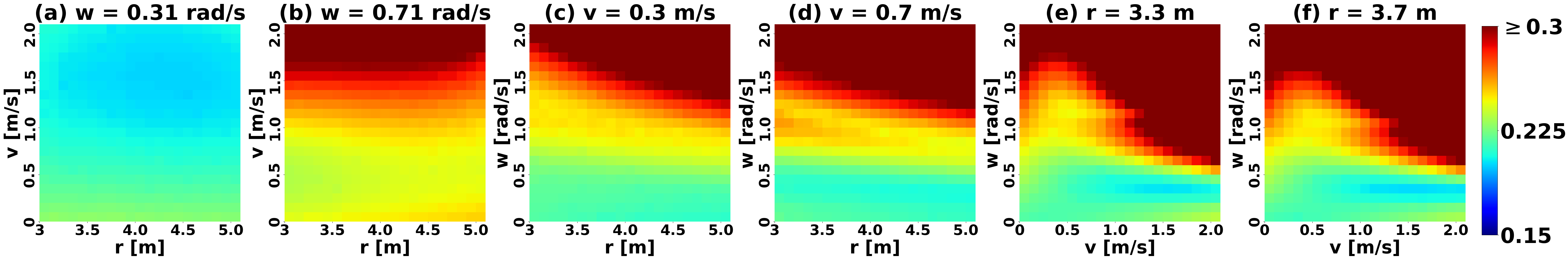}
\caption{Mean errors for tracking a landing trajectory with different $(r,v,\omega)$ settings.
        (a-b) error distribution of $r-v$ with $\omega=$ 0.31 and 0.71. (c-d) error distribution of $\omega-r$ with $v=$ 0.3 and 0.7. (e-f) error distribution of $\omega-v$ with $r=$ 3.3 and 3.7.}
\label{fixed_plan_mean}
\end{figure*}
\begin{figure}[!t]
\centering
\setlength{\belowcaptionskip}{-0.50cm}
\includegraphics[width=0.5\textwidth]{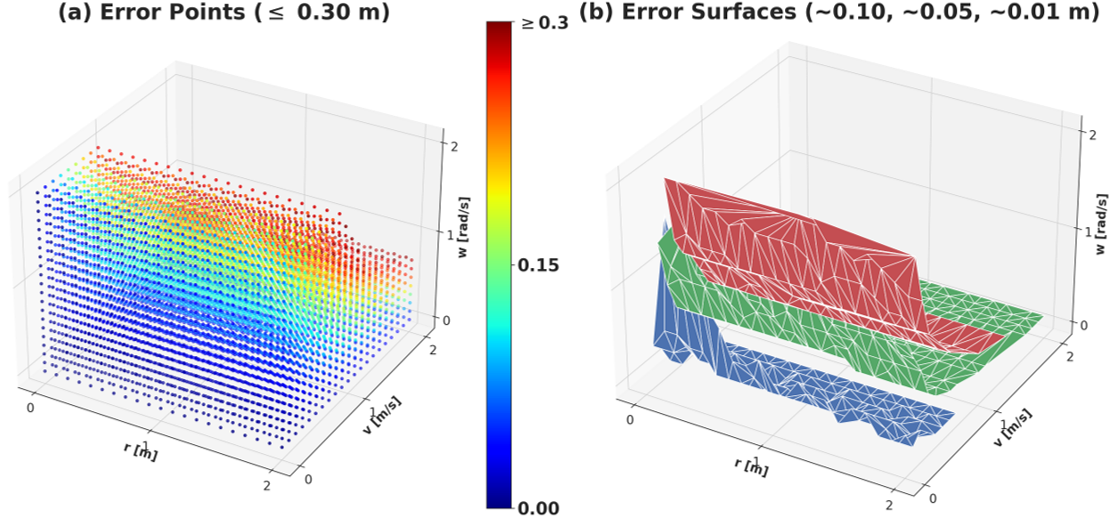}
\caption{ Mean errors for tracking a point with different $(r,v,\omega)$ settings.
        (a) illustrates the data points with tracking error $\leq$ 0.30 m. (b) shows three surfaces of data points with around tracking errors of 0.10 (red), 0.05 (green), and 0.01 (blue) m.  }
\label{fixed_point_scatter}
\end{figure}

Consider a UAV-UGV cooperative system, the uncontrolled motions of the UGV place a crucial role for the system performance.
Most results of UAV-UGV cooperative tasks, such as the autonomous landing in \cite{niu_vision_landing} and \cite{wang_vision_landing}, only show the UGV with constant linear velocities without any aggressive rotational movements. 
However, whether that UAV can track the UGV with both aggressive linear and angular motions well is the key point to consider when applying this system to corresponding tasks. 
From Equ. \ref{eq:v_dot_imu} it can be inferred that ${}^N\boldsymbol{p}_B$, ${}^N\widehat{\boldsymbol{\Omega}}_N$, ${}^N\boldsymbol{\beta}_N$, and  ${}^N\widehat{\boldsymbol{a}}_N$ affect the relative acceleration ${}^N\dot{\boldsymbol{v}}_B$.
Meanwhile, ${}^N\boldsymbol{\Omega}_N$, ${}^N\boldsymbol{p}_B$, and the relative linear velocity in $W$, $\dot{\boldsymbol{t}}_{\sss \uvec{NB}}$, are coupled in the relative velocity (Equ. \ref{pdot}). 
In order to test the performance of the proposed controller, we decouple the variables and pick the parameters $(r, v, \omega)$ to conduct parameter study.
These parameters stand for the range (x-y plane) between robots, linear and angular velocity of the target, which also fits the cooperative task intuitively. 
Parameters definition and theoretical analysis is as following. 
\begin{itemize}
\setlength\itemsep{0.5em}
\item $r \triangleq -{}^N\boldsymbol{p}_B^x$, s.t. ${}^N\boldsymbol{p}_B^x<0$, ${}^N\boldsymbol{p}_B^y=0$, ${}^N\boldsymbol{p}_B^z=z(t)\geq0$ where $z(t)$ can be fixed or time-varying
\item $v \triangleq ({}^N\boldsymbol{R}_W\dot{\boldsymbol{t}}_N)^x$, s.t. $({}^N\boldsymbol{R}_W\dot{\boldsymbol{t}}_N)^x > 0$, $({}^N\boldsymbol{R}_W\dot{\boldsymbol{t}}_N)^y=0$, $({}^N\boldsymbol{R}_W\dot{\boldsymbol{t}}_N)^z=0$
\item $\omega \triangleq {}^N\boldsymbol{\Omega}_N^z$, s.t. ${}^N\boldsymbol{\Omega}_N^x=0$, ${}^N\boldsymbol{\Omega}_N^y=0$, ${}^N\boldsymbol{\Omega}_N^z>0$ 
\end{itemize}
For simplicity, if no other specification, the units of the configuration $(r, v, \omega)$ are m, m/s, and rad/s by default, respectively. 
We expand Equ. \ref{pdot} and \ref{eq:v_dot_imu} in three dimensions to show how $(r, v, \omega)$ are involved in the system model:
\begin{equation}
\begin{aligned}
\label{exp}
{}^N{\boldsymbol{{v}}}_B^x &= ({}^N\boldsymbol{R}_W\dot{\boldsymbol{t}}_B)^x - v \\
{}^N{\boldsymbol{{v}}}_B^y &= ({}^N\boldsymbol{R}_W\dot{\boldsymbol{t}}_B)^y - wr \\
{}^N{\boldsymbol{{v}}}_B^z &= ({}^N\boldsymbol{R}_W\dot{\boldsymbol{t}}_B)^z \\
{}^N\dot{\boldsymbol{{v}}}_B^x &= 2 \omega ({}^N\boldsymbol{R}_W\dot{\boldsymbol{t}}_B)^y + ({}^N\boldsymbol{R}_{B}{}^B\boldsymbol{T}_B)^x - ({}^N\widehat{\boldsymbol{a}}_{N})^x \\
{}^N\dot{\boldsymbol{{v}}}_B^y &= -2 \omega ({}^N\boldsymbol{R}_W\dot{\boldsymbol{t}}_B)^x + 2 \omega v + ({}^N\boldsymbol{R}_{B}{}^B\boldsymbol{T}_B)^y - ({}^N\widehat{\boldsymbol{a}}_{N})^y \\
{}^N\dot{\boldsymbol{{v}}}_B^z &= ({}^N\boldsymbol{R}_{B}{}^B\boldsymbol{T}_B)^z - 9.8
\end{aligned}
\end{equation}

\begin{figure}[!t]
\setlength{\belowcaptionskip}{-0.50cm}
\centering
\includegraphics[width=0.5\textwidth]{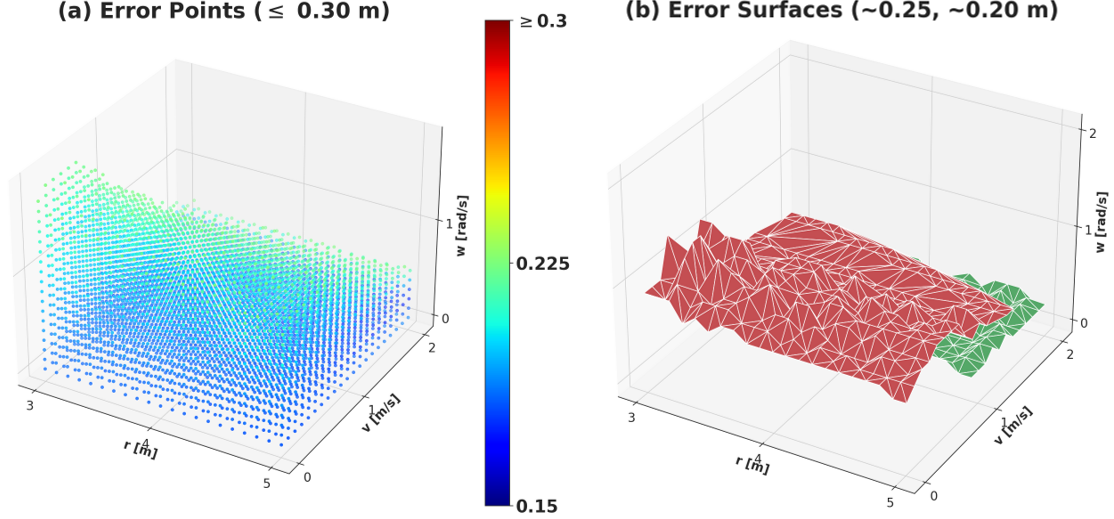}
\caption{ Mean errors for tracking a landing trajectory with different $(r,v,\omega)$ settings.
        (a) illustrates the data points with tracking error $\leq$ 0.30 m. (b) shows three surfaces of data points with around tracking errors of 0.25 (red) and 0.20 (green) m. }
\label{fixed_plan_mean_scatter}
\end{figure}

\vspace{-0.05cm}
In the experiment, apart from the range parameter, the target UGV is programmed with a circular motion with different $v,w$ combination in the world frame (the radius of circle is $v/\omega$). 
We classify the experiments into two schemes, fixed point and fixed plan scheme.
The first is to control the quadrotor follow a fixed point (e.g., ${}^N{\boldsymbol{{p}}}_B = (-r, 0, z)^\top$) in the non-inertial frame. 
We set the $z$ as 2.0 m in simulation and 1.0 m in real experiment.
For example, the configuration $(r,v,\omega)=(\underline{1.0 \text{ m}}, \underline{1.0 \text{ m/s}}, \underline{0.5 \text{ rad/s}})$ represents the quadrotor will follow a fixed point $(\underline{-1.0 \text{ m}}, 0.0 \text{ m}, 2.0 \text{ m})$ in the non-inertial frame with forward \underline{1.0 m/s} speed and counter-clockwise \underline{0.5 rad/s} rotating speed. 
The other test scheme is fixed plan experiment. The configuration has same meaning with the $(v,w)$ but the $r$ stands for the inertial range of the fixed landing trajectory. 
The quadrotor will follow the pre-computed trajectory to land at the origin of the non-inertial frame. 
For example, $\underline{r=1.0}$ represents that the landing trajectory will start at ${}^N{\boldsymbol{{p}}}_B=(\underline{-1.0}, 0.0, 2.0)^\top$ and end at ${}^N{\boldsymbol{{p}}}_B=(0,0,0)^\top$. 
For each scheme, we define the tracking error $e_i$ at each control iteration $i$ as $e_i = \lVert ({}^N\boldsymbol{p}_B)_{\text{est}} - ({}^N\boldsymbol{p}_B)_{\text{ref},i}(0) \rVert$ which is the distance that the the current estimated point deviates from the first point of the reference window.
The mean tracking error for each scheme is defined as $\bar{e} = \sum_i e_i / M$, where $M$ is the total number of iterations.

\vspace{-0.25cm}
\subsection{Simulation}

The numerical simulation is performed on a work station with an AMD Ryzen PRO 5995WX CPU, 
where we use 64 Docker containers simultaneously simulating the quadrotor system model with different parameter configurations using the proposed controller. 
{The MPC in the simulation runs over 100Hz.}
For all simulations, the range of the control inputs is set to
\begin{itemize}
        \item $T \in [2.0, 20.0] \;\text{m}/\text{s}^2$ 
        \item ${}^B\boldsymbol{\Omega}_B^i\in [-3.14, 3.14]$ rad/s $\forall i \in \{x,y,z\}$
\end{itemize}
The penalty matrices $\boldsymbol{Q}$ and $\boldsymbol{R}$ and the constraints are the same. 
The simulated relative estimation is added with a Gaussian noise to represent real sensors, where
$\boldsymbol{\sigma}({{}^N\boldsymbol{p}_B})^i = 0.025 \;\text{m}$, 
$\boldsymbol{\sigma}({{}^N\boldsymbol{v}_B})^i = 0.025 \;\text{m}/\text{s}$, 
$\boldsymbol{\sigma}({\theta({}^N\boldsymbol{q}_B)})^i = 0.044 \;\text{rad}$ ($5^\circ$, $\theta(\cdot)$ is the rotation angle of the quaternion along the rotation axis), 
$\boldsymbol{\sigma}({{}^N\widehat{\boldsymbol{a}}_N})^i = 0.025 \;\text{m}/\text{s}^2$, 
and $\boldsymbol{\sigma}({{}^N\boldsymbol{\Omega}_N})^i = 0.044 \;\text{rad}/\text{s}$ ($5^\circ/s$), 
where $i\in\{x,y,z\}$.

The parameter configurations of the fixed-point scheme are set with
$r \in [0.0, 2,0]$, $v \in [0.0, 2.0]$, and $\omega \in [0.01, 2.01]$ all with stepsizes of $0.1$.  
Fig. \ref{fixed_point} illustrates the tracking errors for different $r, v, \omega$ combinations by selecting two example values for each parameter. 
The parameters for the fixed-plan scheme are: 
$r \in [3.0, 5.0]$, $v \in [0.0, 2.0]$, and $\omega \in [0.01, 2.01]$ all with stepsizes of $0.1$.
Fig. \ref{fixed_plan_mean} illustrates the mean tracking errors with the same logic.
We consider tracking error more than 0.30 m as tracking failure and they are shown in the dark red regions in Fig. \ref{fixed_point} and \ref{fixed_plan_mean}.
In Fig. \ref{fixed_point_scatter} and \ref{fixed_plan_mean_scatter}, we show all the errors in 3D in the left and selected error surfaces (relation between result error and $r, v, \omega$) in the right. 
From these figures, we can conclude that the angular velocity $\omega$ of the non-inertial frame affects the tracking error most.
For a fixed $\omega$, the tracking error increases as $v$ grows.
For a landing task, the simulations can help find the safe $(r,v,\omega)$ parameters that can guarantee the success of landing. 

\vspace{-0.25cm}
\subsection{Real-World Experiment}


\begin{figure}[!t]
\centering
\setlength{\belowcaptionskip}{-0.50cm}
\includegraphics[width=0.5\textwidth]{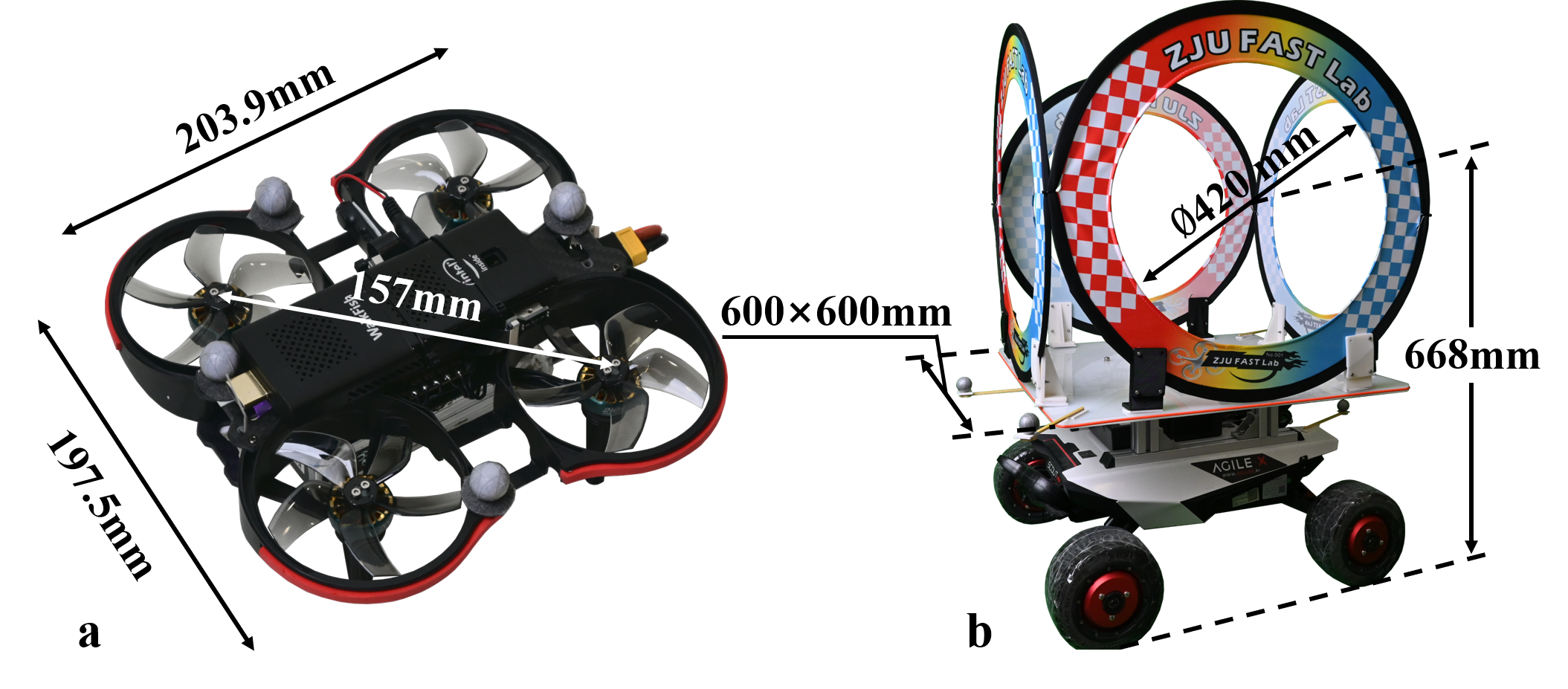}
\caption{The quadrotor and UGV in our real experiments.}
\label{real}
\end{figure}

We have conducted both indoor and outdoor experiments to verify our system.
{The relative estimation can be generated directly from CREPES\cite{CREPES} or computed from NOKOV motion capture system for outdoor or indoor experiments respectively. }
Fig. \ref{real} shows the UAV and UGV platform and their dimensions.
The quadrotor weighs 591.8g and has	a thrust-to-weight ratio of 2.03.
The CoNi-MPC of UAV runs on a onboard computer with an Intel Celeron J4125 processor (2 to 2.7 GHz) and 8GB RAM.
{The proposed CoNi-MPC has computation time (one iteration) of 4.58 ms on average with standard deviation of 1.2 ms, which runs over 100Hz on the onboard computer.}

\begin{figure}[!t]
\centering
\vspace{-0.25cm}
\setlength{\belowcaptionskip}{-0.40cm}
\includegraphics[width=.5\textwidth]{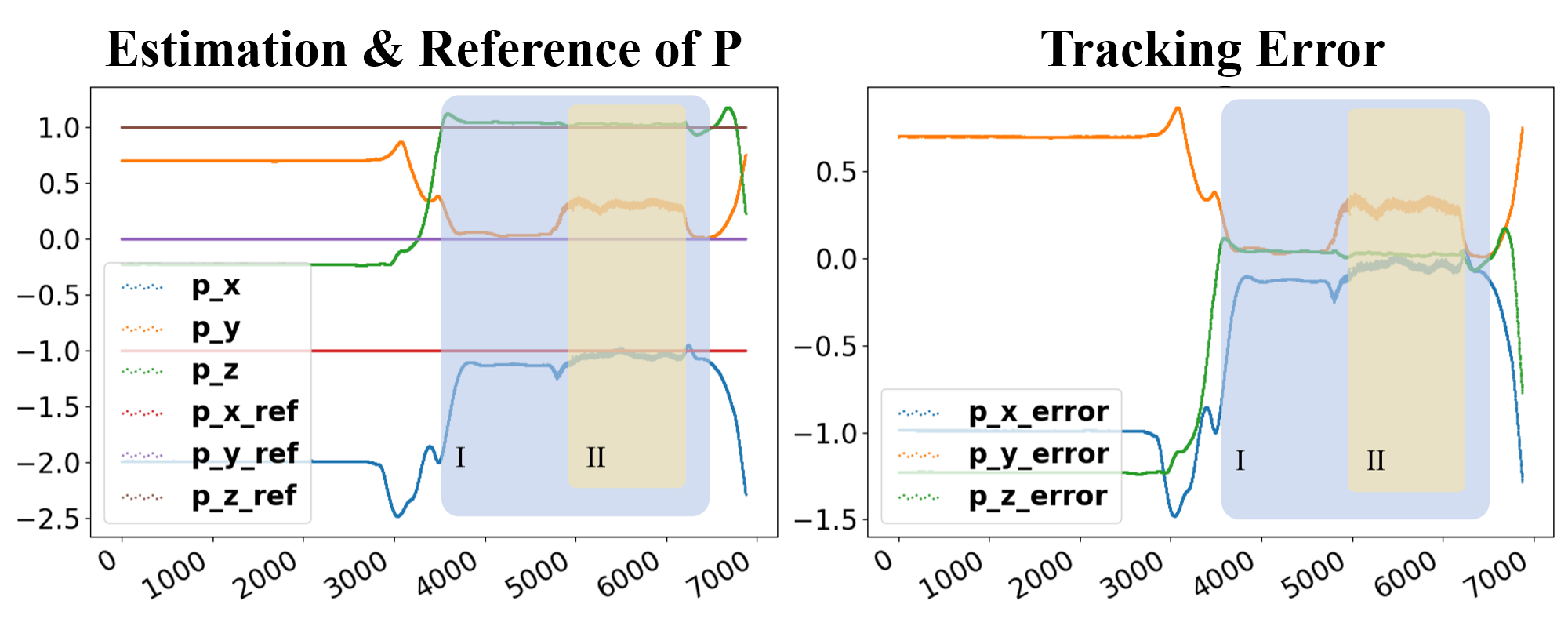}
\caption{Fixed point experiment with $(1.0, 1.0, 0.71)$ setting. Blue (\textbf{I}) is for off-board mode; orange (\textbf{II}) is for UGV moving.}
\label{exp_non_one_point}
\end{figure}

\begin{figure}[!t]
\centering
\setlength{\belowcaptionskip}{-0.40cm}
\includegraphics[width=.5\textwidth]{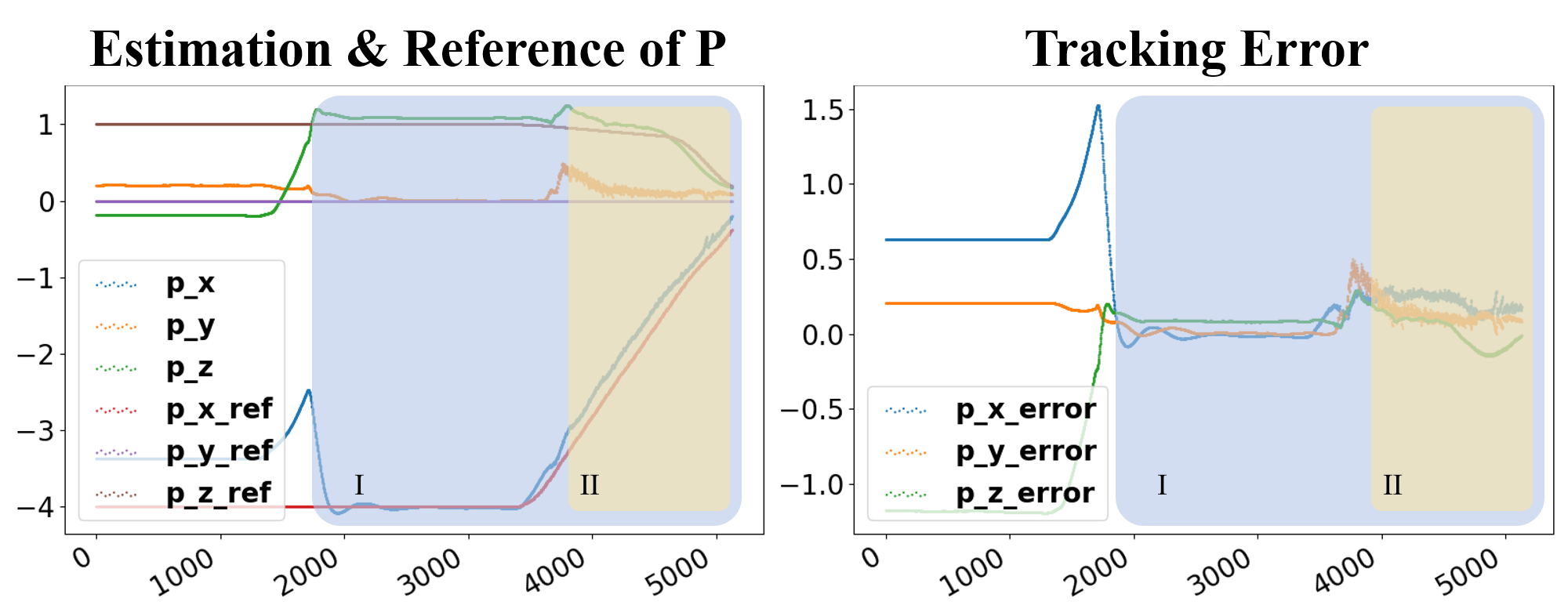}
\caption{Fixed plan experiment with $(4.0, 1.0, 0.71)$ setting. Blue (\textbf{I}) is for off-board mode; orange (\textbf{II}) is for UGV moving.}
\label{exp_non_landing}
\end{figure}

According to the numerical simulation results, we select several representative parameters for both the fixed point and fixed plan schemes.
The constraints in the MPC formulation are the same as that in the simulation.
The parameters and the corresponding average tracking errors of both schemes are listed in Table \ref{avg_error}, 
We plot the tracking performance alone with time of two tests in details in Fig. \ref{exp_non_one_point} and \ref{exp_non_landing}.

\begin{figure}[!t]
\centering
\setlength{\abovecaptionskip}{-0.05cm}
\setlength{\belowcaptionskip}{-0.25cm}
\includegraphics[width=.5\textwidth]{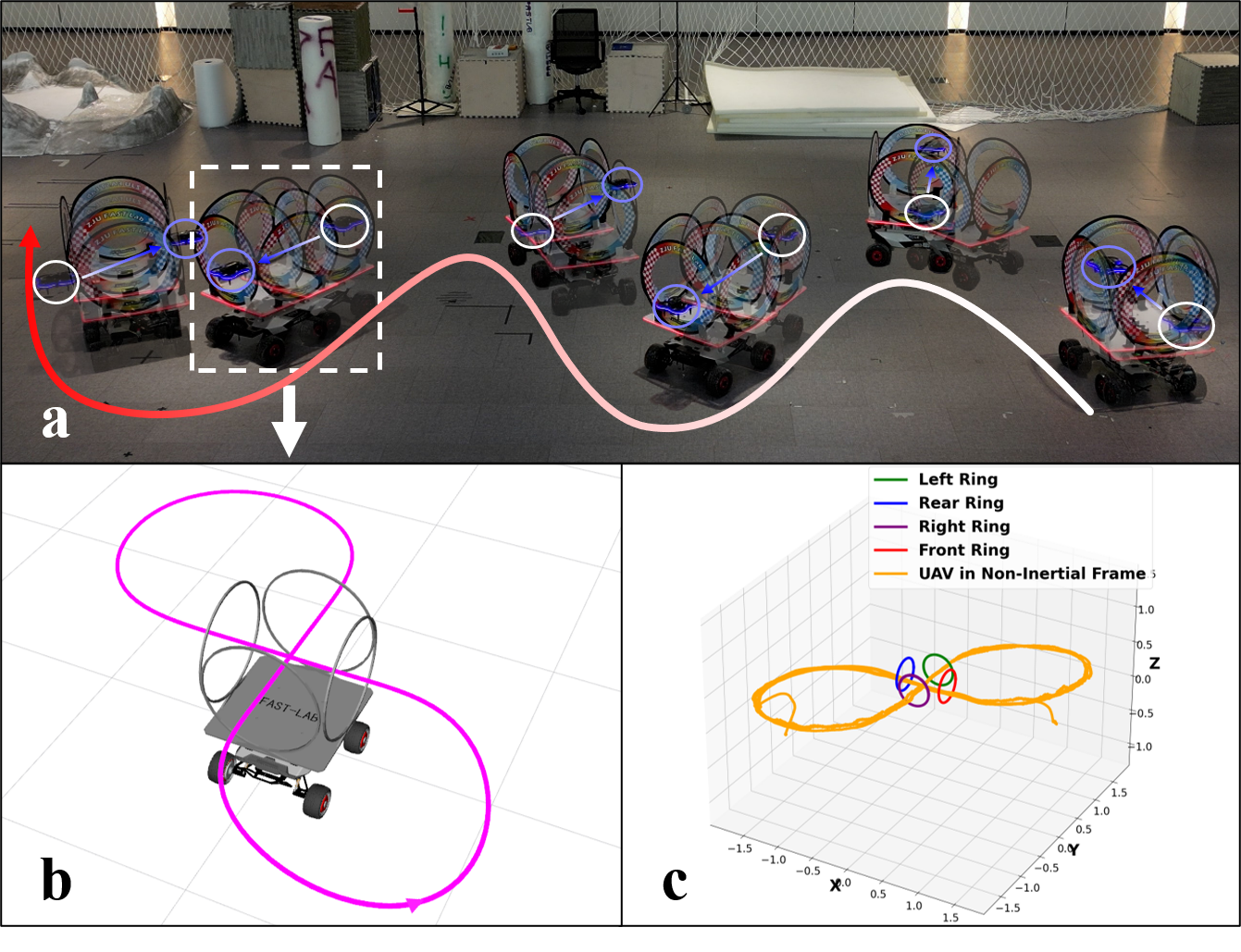}
\caption{Rings crossing experiment. The UGV moves with an average speed of 0.15 m/s and a maximal speed of 0.38 m/s while the UAV can cross the ring attached to it. }
\label{cross_ring}
\end{figure}

\begin{figure}[!t]
\centering
\setlength{\abovecaptionskip}{-0.05cm}
\setlength{\belowcaptionskip}{-0.50cm}
\includegraphics[width=.5\textwidth]{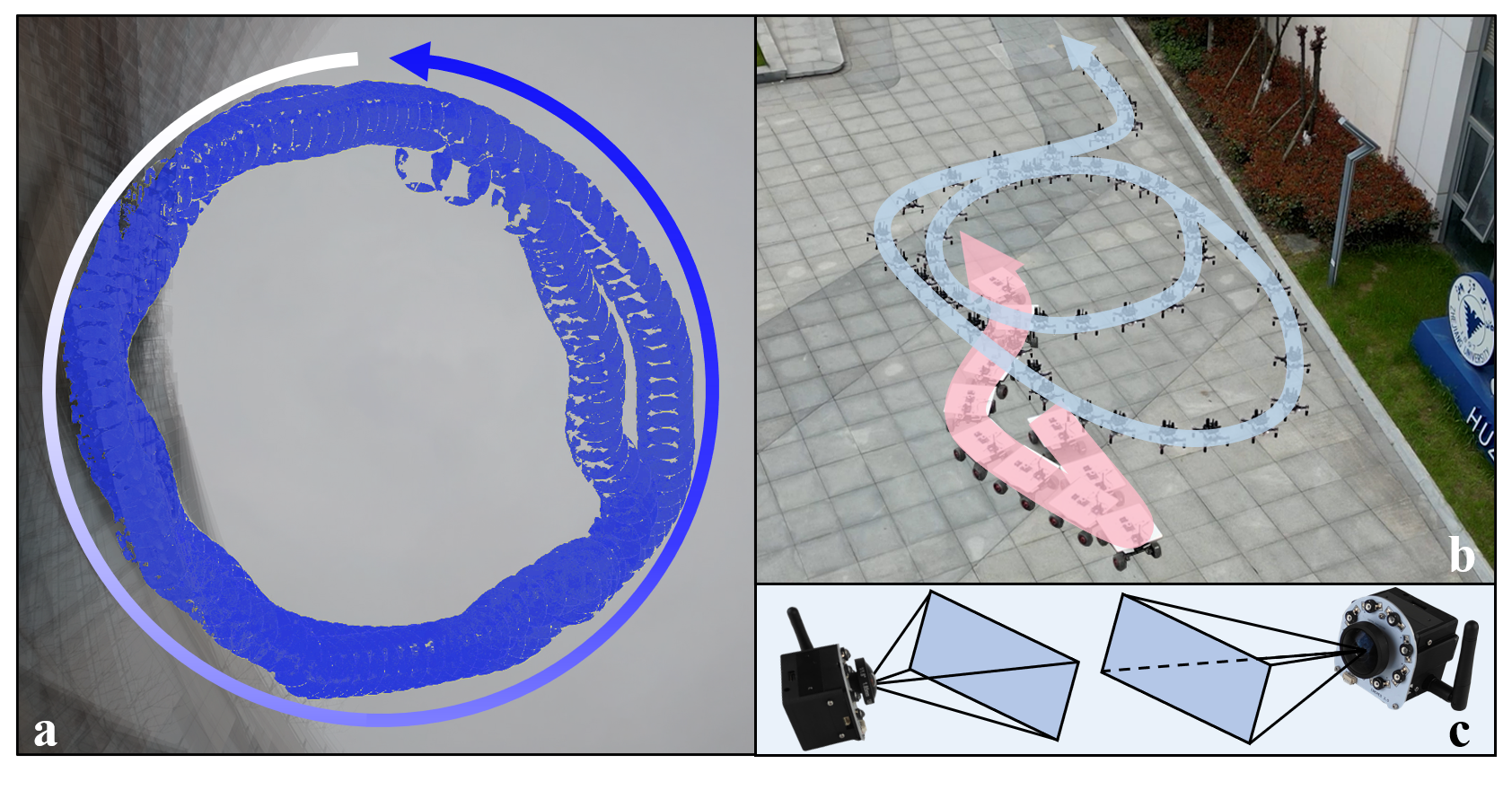}
\caption{Outdoor experiment. (a) is shot from the camera on the UGV, and (b) is shot from third-person view. The UAV follows a circular trajectory while the UGV arbitrarily moves. (c) shows CREPES devices we use in the outdoor experiment. }
\label{outdoor}
\end{figure}

Fig. \ref{cross_ring} shows an qualitative result of a more demanding task, dynamic multi-ring crossing experiment. By simply apply a pre-computed 8-shape trajectory to CoNi-MPC, the quadrotor can continuously cross four rings attached to the four sides of a UGV while the UGV moves along an S-shape trajectory in the world frame. 
The diameter of rings for the UAV to cross is only two times of the UAV's width, about 100 mm free at each side. 
Based on CREPES \cite{CREPES}, we also performed outdoor experiments where the UGV is controlled arbitrarily and the UAV follows a circle trajectory. 
Please note there is no GPS/SLAM/anchors technology used to realize this task. Fig. \ref{outdoor} shows the tracking snapshots. 

\newcolumntype{N}{>{\centering\arraybackslash}m{.1\textwidth}}
\begin{table}[!t]
\caption{Average Tracking Errors of Real Experiments}
\begin{center}
\begin{tabular}{N N | N N}
        \toprule
        \multicolumn{2}{c}{\textbf{Fixed point}} & \multicolumn{2}{c}{\textbf{Fixed plan}} \\
        \cmidrule(lr){1-2}
        \cmidrule(lr){3-4}
        $(r, v, \omega)$ & Error [m] & $(r, v, \omega)$ & Error [m] \\
        \cmidrule(lr){1-1}
        \cmidrule(lr){2-2}
        \cmidrule(lr){3-3}
        \cmidrule(lr){4-4}
        $(1.0,1.0,0.31)$ & 0.13 & ${(\textbf{4.0},\textbf{1.0},\textbf{0.31})}$ & {\textbf{0.15}} \\
        $(1.0,1.0,0.71)$ & 0.23 & $(4.0,1.0,0.71)$ & 0.19 \\
        $(1.0,0.3,1.0)$ & 0.13 & $(4.0,0.3,1.0)$ & 0.21 \\
        $(1.0,0.7,1.0)$ & 0.14 & $(4.0,0.7,1.0)$ & 0.24 \\
        ${(\textbf{0.3},\textbf{1.0},\textbf{1.0})}$ & {\textbf{0.11}} & $(3.3,1.0,1.0)$ & 0.18 \\
        $(0.7,1.0,1.0)$ & 0.16 & $(3.7,1.0,1.0)$ & 0.23 \\
        \bottomrule
\end{tabular}
\label{avg_error}
\end{center}
\end{table}

\vspace{-0.25cm}
\section{Conclusion}
In this work, we present a thorough relative system dynamic model for a quadrotor in a non-inertial frame with linear and angular motions.
Based on this model, we design and implement CoNi-MPC targeting cooperative UAV-UGV cooperation tasks, by taking the UGV as a non-inertial frame.  
Unlike traditional methods, this method bypasses the dependency of global state estimation of the agent and/or the target in the world frame. 
The system also avoids relying on prior knowledge of the target and does not need complex trajectory re-planning. 
CoNi-MPC only requires the relative states (pose and velocity), the angular velocity and the accelerations of the target, which can be obtained by relative localization methods and ubiquitous MEMS IMU sensors, respectively. 
We have performed extensive fixed-point and fixed-plan simulations and considerable real world experiments to test the proposed system.
Experiment results show that the controller has promising robustness and tracking performance.
For future works, this method can be extended to achieve multi-agent formation control and more demanding cooperative tasks.
\vspace{-0.25cm}

\bibliographystyle{IEEEtran}
\bibliography{paper}

\end{document}